\documentclass[letterpaper]{article} 
\usepackage{aaai2026}  
\usepackage{times}  
\usepackage{helvet}  
\usepackage{courier}  
\usepackage[hyphens]{url}  
\usepackage{graphicx} 
\usepackage{xcolor}
\usepackage{natbib}  
\usepackage{caption} 
\usepackage{xcolor}
\usepackage{array}
\usepackage{longtable}
\usepackage{caption}
\usepackage{amsmath}
\usepackage[table]{xcolor}
\usepackage[utf8]{inputenc} 
\usepackage[T1]{fontenc}    
\usepackage{url}            
\usepackage{booktabs}       
\usepackage{amsfonts}       
\usepackage{nicefrac}       
\usepackage{microtype}      
\usepackage{xcolor}         
\usepackage{graphicx}
\usepackage{tikz}
\usepackage{multicol}
\usepackage{multirow}
\usepackage{mathrsfs}

\usepackage{algpseudocode}
\usepackage{algorithm}  
\usepackage{algorithmicx}  
\usepackage{tabularx}
\usepackage{subcaption}

\renewcommand{\vec}[1]{\mathbf{#1}}

\usepackage{newfloat}
\usepackage{listings}
\DeclareCaptionStyle{ruled}{labelfont=normalfont,labelsep=colon,strut=off} 
\floatstyle{ruled}
\newfloat{listing}{tb}{lst}{}
\floatname{listing}{Listing}
\title{Hyperbolic Hierarchical Alignment Reasoning Network for Text-3D Retrieval}
\author{
    Wenrui Li\equalcontrib\textsuperscript{\rm 1}, Yidan Lu\equalcontrib\textsuperscript{\rm 1}, Yeyu Chai\textsuperscript{\rm 1}, Rui Zhao\textsuperscript{\rm 2}, Hengyu Man\textsuperscript{\rm 1} and Xiaopeng Fan\textsuperscript{\rm 1}\textsuperscript{\rm 3}\textsuperscript{\rm 4}\thanks{Corresponding author.}\\
}
\affiliations{
    \textsuperscript{\rm 1}Harbin Institute of Technology \textsuperscript{\rm 2}Nanyang Technological University\\ \textsuperscript{\rm 3}Harbin Institute of Technology Suzhou Research Institute \textsuperscript{\rm 4} Peng Cheng Laboratory\\
    liwr618@163.com; 24S103311@stu.hit.edu.cn; 23s136130@stu.hit.edu.cn\\zhao.rui@ntu.edu.sg; manhengyu@hotmail.com; fxp@hit.edu.cn}

\usepackage{bibentry}

\begin{document}

\maketitle

\begin{abstract}
With the daily influx of 3D data on the internet, text-3D retrieval has gained increasing attention. However, current methods face two major challenges: Hierarchy Representation Collapse (HRC) and Redundancy-Induced Saliency Dilution (RISD). HRC compresses abstract-to-specific and whole-to-part hierarchies in Euclidean embeddings, while RISD averages noisy fragments, obscuring critical semantic cues and diminishing the model’s ability to distinguish hard negatives. To address these challenges, we introduce the Hyperbolic Hierarchical Alignment Reasoning Network (H$^{2}$ARN) for text-3D retrieval. H$^{2}$ARN embeds both text and 3D data in a Lorentz-model hyperbolic space, where exponential volume growth inherently preserves hierarchical distances. A hierarchical ordering loss constructs a shrinking entailment cone around each text vector, ensuring that the matched 3D instance falls within the cone, while an instance-level contrastive loss jointly enforces separation from non-matching samples. To tackle RISD, we propose a contribution-aware hyperbolic aggregation module that leverages Lorentzian distance to assess the relevance of each local feature and applies contribution-weighted aggregation guided by hyperbolic geometry, enhancing discriminative regions while suppressing redundancy without additional supervision. We also release the expanded T3DR-HIT v2 benchmark, which contains 8,935 text-to-3D pairs, 2.6 times the original size, covering both fine-grained cultural artefacts and complex indoor scenes. Our codes are available at \url{https://github.com/liwrui/H2ARN}.
\end{abstract}


\section{Introduction}
\begin{figure}[t]
    \centering
    \includegraphics[width=\columnwidth]{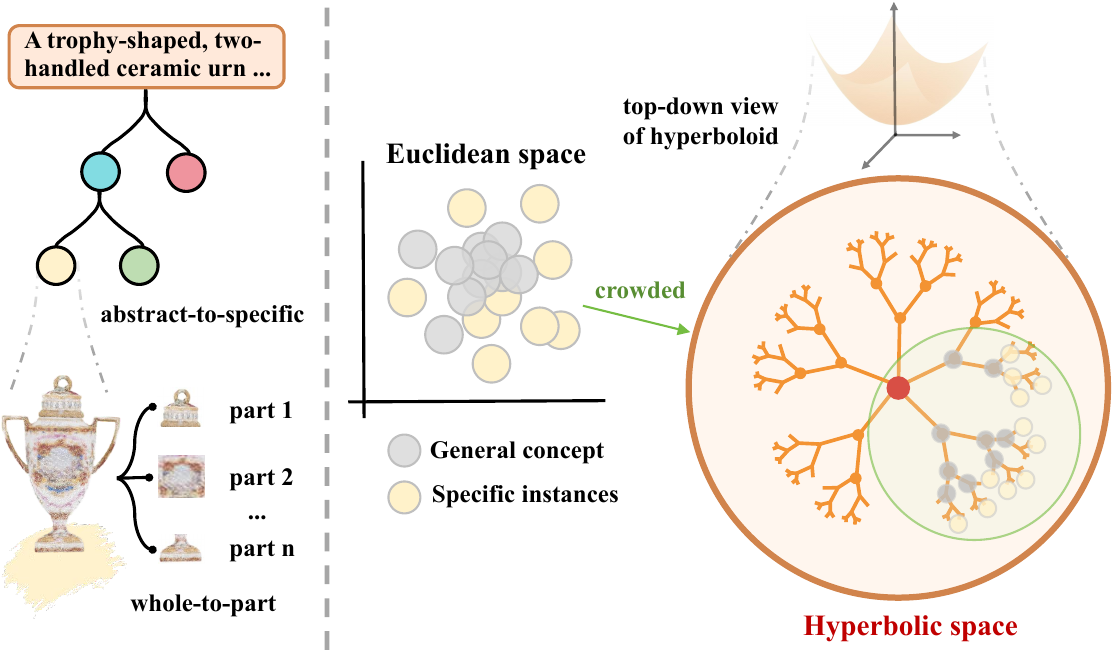}
    \caption{Conceptual illustration of hierarchical data representation. \textbf{Left:} The exponentially growing tree structures inherent in both abstract-to-specific semantics and whole-to-part geometry. \textbf{Right:} Comparison of embedding spaces. Euclidean space suffers from a "crowding" effect, whereas hyperbolic space naturally preserves the hierarchy. In hyperbolic space, the origin represents the most general concepts, with distance from the origin encoding semantic specificity.}
    \label{fig:figure_1}
\end{figure}

With the rapid increase in the volume and variety of 3D data available online, text-3D retrieval has attracted significant attention for its broad application potential. Unlike traditional cross-modal retrieval tasks limited to 2D alignment, this area directly maps natural language to rich geometric, topological, and textural information. This capability enables more accurate and actionable content analysis, benefiting applications such as  3D crack recognition \cite{Chen_01,Chen_02,Chen_03}, and multimodal processing \cite{bai01,Bai02,xiao01,xiao02,zhang01,zhang02,zhuoyuan01,wenrui111,wenrui222,wenrui333,wenrui444,bao01,bao02}.

However, bridging the semantic gap between language and 3D geometry presents substantially greater challenges than traditional cross-modal retrieval tasks \cite{li2023style}. Both 3D data and natural language exhibit inherent tree-like hierarchies: semantics evolve from abstract concepts to concrete details, while geometry transitions from holistic structures to fine-grained components. This hierarchical structure leads to an exponential growth in the number of nodes with increasing depth. When embedded into Euclidean or conventional Riemannian spaces, which grow at most polynomially with radius \cite{lee2018introduction}, a “crowding” effect becomes inevitable, as shown in the right panel of Figure~\ref{fig:figure_1}. Samples that are semantically distinct but structurally similar are compressed into close proximity in high-dimensional embedding spaces. Moreover, real-world 3D data often contain artifacts and texture noise that introduce unavoidable redundancy. Mainstream methods typically use mean pooling to aggregate local fragments into global representations, assuming equal contribution from all parts. As a result, crucial geometric features are often diluted by semantically irrelevant noise. Therefore, the existing text-3D retrieval methods facing two fundamental challenges: Hierarchy Representation Collapse (HRC) and Redundancy-Induced Saliency Dilution (RISD).

First, HRC disrupts semantic consistency across modalities. Both natural language and 3D geometry inherently follow a tree-structured hierarchy, progressing from abstract to concrete semantics and from global to local geometric structures \cite{li2023uni3dl}. The number of nodes increases exponentially with hierarchical depth. For example, as illustrated in the left panel of Figure~\ref{fig:figure_1}, the text “A trophy-shaped, two-handled ceramic urn” denotes an abstract concept, while its corresponding 3D point cloud encodes fine-grained details such as the ornate handles and specific surface patterns. These details correspond to more specific instances within the same hierarchy. When this exponentially growing hierarchy is projected into Euclidean or conventional Riemannian spaces \cite{li2025riemann}, which expand at most polynomially with radius, intrinsic tree distances become severely distorted. General concepts and specific instances are compressed into crowded regions, resulting in overlap and ambiguity within the embedding space. As a result, the model fails to preserve the intrinsic property that higher-level concepts should encompass more neighbors, whereas lower-level instances should remain distinguishable. This distortion contributes to simultaneous declines in both recall and precision.

Second, RISD amplifies discriminative errors during feature aggregation. Real-world 3D data often contain redundant fragments, such as scanning artifacts and decorative textures, while natural language descriptions may include non-discriminative elements, such as prepositions and function words. When equal-weighted strategies such as mean pooling are employed, the contribution of local fragments to the global representation is uniformly smoothed. Consequently, critical geometric and semantic cues are averaged out, diminishing the embedding’s ability to distinguish hard negative samples.

To address these challenges, we propose the Hyperbolic Hierarchical Alignment Reasoning Network (H$^{2}$ARN), which jointly embeds textual descriptions and 3D point clouds into a Lorentzian hyperbolic space with constant negative curvature. Due to its exponential volume growth with respect to radius \cite{gromov1987hyperbolic}, this space naturally accommodates tree-structured hierarchies (Figure~\ref{fig:figure_1}, right panel). The origin represents the most general concepts, and embeddings positioned closer to the origin carry more abstract semantics, which inherently subsume the more specific instances situated farther away. Leveraging this geometric framework, we first impose a cross-modal constraint: text embeddings are required to lie closer to the origin than their corresponding point cloud embeddings, thereby spatially encoding the abstract-to-specific relationship. Next, we introduce a contribution-aware intra-modal aggregation mechanism. Local geometric features and word tokens are treated as leaf nodes and are contextually enriched via a self-attention module. These enriched representations, along with an initial global anchor obtained through mean pooling, are projected into hyperbolic space. The Lorentzian distance is then used to quantify each leaf node’s semantic contribution to the anchor, assigning it an importance weight. This guides a weighted aggregation process that produces a final global representation, semantically cleaner and positioned deeper within the hierarchy. The optimization process incorporates two geometric loss functions. The first is a Lorentzian contrastive loss, which promotes instance-level alignment by pulling matched pairs closer and pushing mismatched pairs farther apart in hyperbolic space. The second is the Hierarchical Ordering Loss, which explicitly encodes the partial ordering of “text entails 3D” through entailment cones. Given a text embedding $x$ and a point cloud embedding $y$, we construct a hyperbolic cone centered at $x$ with a radially shrinking aperture. If $y$ lies within the cone, the partial order is satisfied and no penalty is incurred. Otherwise, the loss is proportional to the angular deviation from the cone boundary. This mechanism dynamically adjusts the cone to encompass relevant 3D instances while excluding unrelated ones during training. The main contributions of this paper are summarized as follows:
\begin{itemize}
    \item We propose the H$^{2}$ARN, which constructs a Lorentzian hyperbolic space with constant negative curvature and introduces a hierarchical ordering loss. By explicitly enforcing partial order constraints via entailment cones in the embedding space, our model effectively mitigates the problem of hierarchical representation collapse.
    \item We introduce a contribution-aware hyperbolic aggregation mechanism that leverages Lorentzian distance to estimate the semantic contribution of each local fragment. When jointly trained with hierarchical ordering loss, the model improves its ability to distinguish hard negative samples without 
    additional supervision. 
    \item We expand the T3DR-HIT dataset to 2.6 times its original size, increasing the number of text–3D pairs from 3,380 to 8,935. Our model demonstrates superior performance and generalization capabilities on the enlarged dataset. We believe that releasing this expanded dataset will benefit the broader research community.
\end{itemize}

\section{Related Work}
\subsection{Cross-modal Retrieval}
Cross-modal retrieval aims to bridge the semantic gap between different data modalities, with feature alignment being the central challenge \cite{haocheng01,yangzhe1,yangzhe2,yangzhe3}. The most prominent subfield is image-text retrieval, where alignment strategies are often categorized into coarse-grained, fine-grained, and hybrid-grained approaches. Coarse-grained methods typically map entire images and sentences to a shared embedding space for holistic comparison \cite{faghri2017vse++}. Some enhance global representations by modeling intra-modal relationships using graph convolutional networks \cite{li2019visual} or transformers \cite{messina2021transformer, li2022image, li2023style}, while others focus on improving the loss functions with instance-level constraints \cite{zheng2020dual}, hierarchical relation modeling \cite{fu2023learning}, or adversarial learning \cite{peng2019cm}. Fine-grained methods, in contrast, focus on aligning local features. A seminal work introduced cross-attention to discover latent alignments between visual and textual fragments \cite{lee2018stacked}. Subsequent approaches refined this core idea by extending alignment to the relation-level \cite{wei2020multi, messina2021fine}, selectively attending to salient local fragments based on global context \cite{zhang2020context, bao2023multi, wang2025global}, or introducing innovative techniques such as multi-level contrastive learning \cite{wu2019unified} and probabilistic modeling \cite{li2024multi}. Hybrid-grained methods unify the strengths of both via strategies such as inferring more accurate matching scores through similarity attention filtering \cite{diao2021similarity}, integrating coarse and fine-grained learning into a unified framework with a consistency-constrained contrastive loss \cite{liu2023efficient}, or employing intra-modal fusion guidance and inter-modal bidirectional guidance \cite{chen2024integrating}.

Inspired by these advances, research has extended to the 3D domain, initially focusing on text-3D shape retrieval. This sub-task focuses on matching textual descriptions with isolated 3D shapes, typically generated models. A pioneering study first achieved end-to-end text-shape retrieval by combining association learning with metric learning \cite{chen2018text2shape}. Subsequent works have largely relied on multi-view renderings, representing 3D shapes as view sequences \cite{han2019y2seq2seq}, learning a trimodal embedding space \cite{ruan2024tricolo}, or using self-and-cross-attention to aggregate multi-view and point cloud features \cite{lin2024sca}, often supplemented by hard negative mining strategies \cite{wu2024com3d}. To move beyond reliance on 2D renderings, some methods have introduced direct matching between 3D shape parts and words \cite{tang2023parts2words} or proposed unified query transformers for joint understanding \cite{li2023uni3dl}. However, by focusing on isolated objects lacking scene details and often neglecting the underlying data distribution, these methods can be susceptible to learning biases.

Different from previous text-3D benchmarks targeted specific scenarios like synthetic indoor scenes \cite{yu2024towards}, a seminal study recently advanced the field by introducing T3DR-HIT, the first large-scale benchmark based on real-world scans of both coarse-grained scenes and fine-grained artifacts \cite{li2025riemann}. The study also introduced a Riemannian attention mechanism to enhance retrieval accuracy. Despite these advances, prior methods primarily operate in Euclidean or conventional Riemannian spaces with limited volume growth, making them susceptible to HRC. Concurrently, their common use of equal-weighted aggregation strategies fails to effectively address the RISD problem. Our work directly addresses these gaps by leveraging the exponential capacity of Lorentzian hyperbolic geometry and a contribution-aware aggregation mechanism, thereby enabling more robust and precise text-3D retrieval.

\section{Method}
\begin{figure*}[t]
    \centering
    \includegraphics[width=\textwidth]{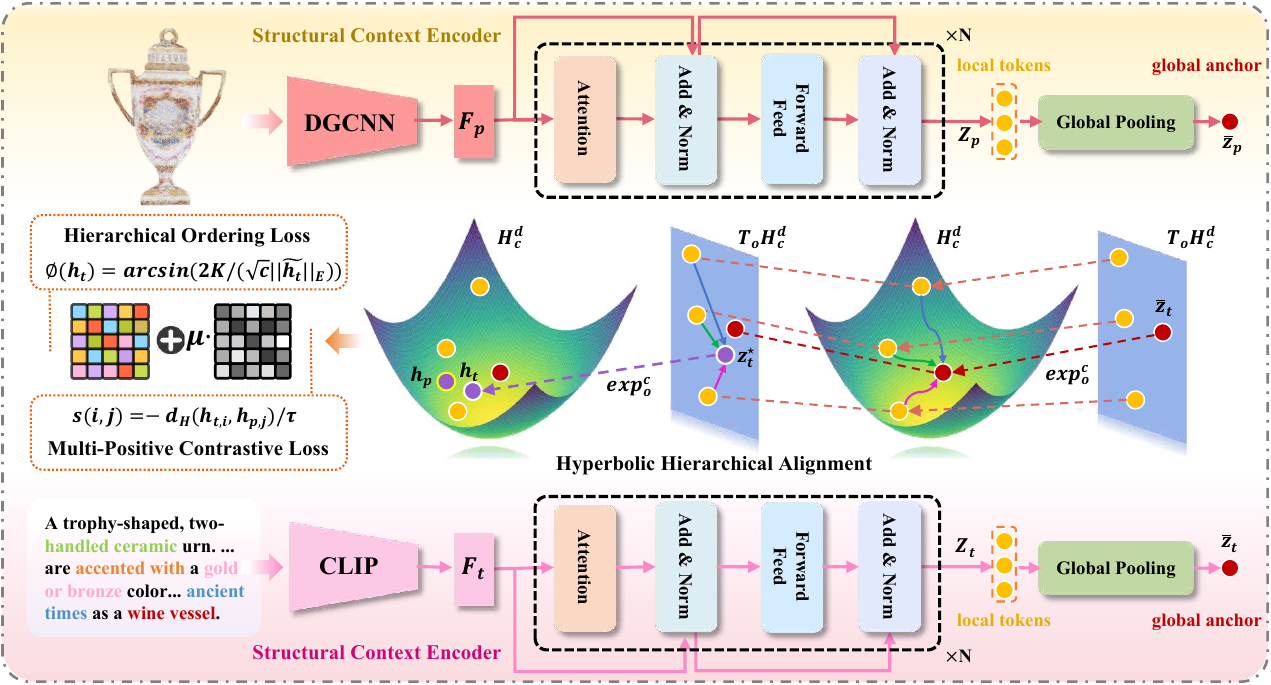}
    \caption{An overview of the H$^{2}$ARN architecture. The Structural Context Encoder first refines local features from each modality in Euclidean space to produce context-aware representations. Subsequently, the Hyperbolic Hierarchical Alignment Module aligns the features in hyperbolic space via a contribution-aware aggregation mechanism and a dual geometric loss, preserving their semantic hierarchy.}
    \label{fig:figure_2}
\end{figure*}

In this section, we present in detail the modeling architecture and learning objectives of \textbf{H$^{2}$ARN}. As illustrated in Figure~\ref{fig:figure_2}, the H$^{2}$ARN framework is composed of two primary modules: a \textbf{Structural Context Encoder} that refines local features in Euclidean space, and a \textbf{Hyperbolic Hierarchical Alignment Module} that embeds and aligns them in hyperbolic space. We first introduce the preliminaries of the Lorentz model of hyperbolic geometry. Then, we discuss the architectural components of our model and its optimization strategy.

\subsection{Preliminaries of Lorentz Model}
Hyperbolic space $\mathbb{H}^{d}$ is a Riemannian manifold with constant negative curvature. Its volume grows exponentially with geodesic radius, mirroring the branching of tree-structured data and easing the crowding that arises in Euclidean embeddings. Several coordinate models represent $\mathbb{H}^{d}$, for example the Poincaré ball or the upper half-space, but these embed the manifold in $\mathbb{R}^{d}$ at the cost of metric distortion. We adopt the Lorentz model because it provides an isometric embedding in the $(d+1)$-dimensional Minkowski space $\mathbb{R}^{1,d}$, preserving distances exactly and enabling stable closed-form geodesic operations. For curvature $-c<0$, the Lorentz model is the future sheet of the two-sheeted hyperboloid in $\mathbb{R}^{1,d}$:
\begin{equation}
\mathbb{H}_{c}^{d} = \{ \vec{u} \in \mathbb{R}^{d+1} : \langle \vec{u}, \vec{u} \rangle_{\mathcal{L}} = -\frac{1}{c}, u_{d+1} > 0 \},
\end{equation}
where $\langle\cdot,\cdot\rangle_{\mathcal L}$ denotes the Lorentz inner product.
For two vectors $\mathbf u,\mathbf v\in\mathbb R^{d+1}$ with spatial parts $\tilde{\mathbf u},\tilde{\mathbf v}\in\mathbb R^{d}$ and time components $u_{d+1},v_{d+1}\in\mathbb R$, the inner product is defined as $\langle \vec{u}, \vec{v} \rangle_{\mathcal{L}} = \langle \tilde{\vec{u}}, \tilde{\vec{v}} \rangle_{E} - u_{d+1}v_{d+1}$, where $\langle \cdot, \cdot \rangle_{E}$ is the standard Euclidean inner product. All vectors on this manifold satisfy the constraint $u_{d+1} = \sqrt{1/c + \|\tilde{\vec{u}}\|_{E}^2}$. Based on the structure, we use the Lorentzian distance to measure the geometric distance between embedded points. This distance corresponds to the length of the shortest path (geodesic) on the manifold, effectively capturing hierarchical semantic relationships. For any two points $\vec{u}, \vec{v} \in \mathbb{H}_c^d$, their Lorentzian distance is defined as:
\begin{equation}
d_{\mathbb{H}}(\vec{u}, \vec{v}) = \frac{1}{\sqrt{c}} \operatorname{arccosh}(-c \langle \vec{u}, \vec{v} \rangle_{\mathcal{L}}).
\end{equation}

To enable feature projection and optimization on the manifold, it is necessary to introduce the tangent space and its mapping to the manifold. The tangent space at a point $\vec{w} \in \mathbb{H}_{c}^{d}$ is a $d$-dimensional Euclidean space that is orthogonal to $\vec{w}$ under the Lorentz inner product:
\begin{equation}
T_{\vec{w}}\mathbb{H}_{c}^{d} = \{ \vec{v} \in \mathbb{R}^{d+1} : \langle \vec{v}, \vec{w} \rangle_{\mathcal{L}} = 0 \},
\end{equation}
where $T_{\vec{w}}\mathbb{H}_{c}^{d}$ denotes the tangent space at point $\vec{w}$.
The exponential map, $\operatorname{exp}_{\vec{w}}^{c}: T_{\vec{w}}\mathbb{H}_{c}^{d} \to \mathbb{H}_{c}^{d}$, serves as a crucial bridge, projecting a vector from the flat tangent space onto the curved manifold. For a general point $\vec{w}$ and a tangent vector $\vec{v} \in T_{\vec{w}}\mathbb{H}_{c}^{d}$, this map is defined as:
\begin{equation}
\operatorname{exp}_{\vec{w}}^{c}(\vec{v}) = \cosh(\sqrt{c}\|\vec{v}\|_{\mathcal{L}})\vec{w} + \frac{\sinh(\sqrt{c}\|\vec{v}\|_{\mathcal{L}})}{\sqrt{c}\|\vec{v}\|_{\mathcal{L}}}\vec{v},
\end{equation}
where $\|\vec{v}\|_{\mathcal{L}} = \sqrt{|\langle \vec{v}, \vec{v} \rangle_{\mathcal{L}}|}$ is the Lorentzian norm.

The exponential map serves as the bridge for lifting Euclidean features into the hyperbolic manifold. In practice, we focus on the exponential map centered at the hyperboloid origin $\mathbf{o}=(0,\dots,0,1/\sqrt{c})$, since any Euclidean feature vector $\mathbf{v}\in\mathbb{R}^d$ lies in the tangent space $T_{\vec{o}}\mathbb{H}_{c}^{d}$. Its temporal component is zero, ensuring that $\langle \vec{o}, \vec{v} \rangle_{\mathcal{L}} = 0$ holds automatically. Substituting $\mathbf{w}=\mathbf{o}$ into the general Lorentz exponential map yields the embedded point $\mathbf{u}=\exp^{c}_{\mathbf{o}}(\mathbf{v})$, where the spatial component is $\mathbf{v}$ scaled by a hyperbolic factor and the time coordinate is determined by curvature. This faithfully positions the feature on the constant-curvature manifold and facilitates geometry-aware learning, which can be written as:
\begin{equation}
\tilde{\vec{u}} = \frac{\sinh(\sqrt{c}\|\vec{v}\|_{E})}{\sqrt{c}\|\vec{v}\|_{E}}\vec{v}, \quad \quad u_{d+1} = \frac{\cosh(\sqrt{c}\|\vec{v}\|_{E})}{\sqrt{c}},
\end{equation}
where the time component $u_{d+1}$ can be equivalently derived from the spatial component via the hyperboloid constraint.

\subsection{Structural Context Encoder}
To capture each modality’s intrinsic structure and furnish context-aware representations, we first feed raw inputs to strong modality-specific backbones, yielding local feature sequences $F_t \in \mathbb{R}^{L_t \times D_{t}}$ for text and $F_p \in \mathbb{R}^{L_p \times D_{p}}$ for point clouds. Text tokens are encoded with CLIP \cite{radford2021learning}, whose large-scale vision–language pre-training supplies rich semantic priors for nuanced descriptions. Point clouds are processed by DGCNN \cite{wang2019dynamic}, which dynamically constructs neighbourhood graphs to model point–point relations while integrating colour channels to preserve fine visual detail. Although these encoders excel at local pattern extraction, the resulting sequences lack global awareness of long-range dependencies. We therefore project the features to a shared latent dimension $d$ and pass them through a stack of Pre-Layer-Norm Transformer blocks. In each block, the sequence is linearly mapped to query, key and value tensors, followed by multi-head scaled dot-product attention that captures cross-token interactions; position-wise feed-forward networks and residual connections then refine the context. This self-attention pipeline endows both modalities with coherent, context-enhanced embeddings that are ready for subsequent hyperbolic alignment.

\subsection{Hyperbolic Hierarchical Alignment Module}

This module is the core of our model, specifically designed to address the dual challenges of HRC and RISD. It achieves this through a novel Contribution-Aware Hyperbolic Aggregation mechanism and a Dual Geometric Loss Function that is computed entirely within the hyperbolic geometry.

\noindent\textbf{Contribution-Aware Hyperbolic Aggregation.} 
The contribution-aware hyperbolic aggregation mechanism addresses RISD by weighting local tokens based on their geometric relevance to global semantics. Let $Z_{t}$ and $Z_{p}$ denote the context-enhanced Euclidean token matrices from the previous stage, with each row vector $z_{i}$ regarded as a leaf node. We first compute an initial anchor $ \bar z = \frac{1}{L}\sum_{i=1}^{L} z_{i}$ and map both the anchor and all leaf nodes to $\mathbb H^{d}_{c}$ using the exponential map defined in Eq. (5). For each leaf, we calculate its Lorentzian distance to the anchor as defined in Eq. (2). A softmax over the negative distances yields contribution weights $\omega_{i}$ that reflect semantic saliency. These weights guide the weighted Euclidean sum $z^{\star} = \sum_{i=1}^{L}\omega_{i} z_{i}$. Since $z^{\star}$ has a smaller norm than any individual $z_{i}$, its hyperbolic image naturally lies closer to the origin, capturing a more abstract and denoised global concept.  The resulting root embeddings, $\mathbf{h}_{t} = \exp^{c}_{\mathbf{o}}(z^{\star}_{t})$ and $\mathbf{h}_{p} = \exp^{c}_{\mathbf{o}}(z^{\star}_{p})$, provide semantically purified representations for text and point clouds, effectively countering the dilution of key cues by redundant fragments. To prevent numerical overflow in the exponential map, each local feature matrix $Z \in \mathbb{R}^{L \times d}$ is scaled by a learnable, modality-specific factor $\alpha$, i.e., $Z' = \alpha Z$, prior to aggregation. 

\noindent\textbf{Dual Geometric Loss Function.} Our optimization objective structures the embedding space to simultaneously ensure instance-level discrimination and preserve the inter-modal abstract-to-specific semantic hierarchy. This is achieved through two synergistic components: a multi-positive contrastive loss constructed from negative Lorentzian distance, which aligns instances by pulling positive pairs closer and pushing negative pairs apart, and a hierarchical ordering loss, which explicitly enforces the "text entails 3D" partial order using entailment cones with a radially narrowing scope.

\noindent \textbf{The Multi-Positive Contrastive Loss.} The multi-positive contrastive loss $\mathcal{L}_{\text{cont}}$ operates at the instance level. We define the similarity score $s(i,j)$ between the final global root embeddings of a text instance $i$ ($\vec{h}_{t,i}$) and a 3D instance $j$ ($\vec{h}_{p,j}$) as their negative Lorentzian distance:
\begin{equation}
s(i,j) = -d_{\mathbb{H}}(\vec{h}_{t,i}, \vec{h}_{p,j}) / \tau,
\end{equation}
where $\tau$ is a temperature hyperparameter.
Based on this similarity, we construct a symmetric InfoNCE-style loss adapted for scenarios where multiple positive samples may exist for a given query in a batch. The loss from text to point cloud, $\mathcal{L}_{t \to p}$, is formulated as:
\begin{equation}
\mathcal{L}_{t \to p} = - \frac{1}{B} \sum_{i=1}^{B} \log \frac{\sum_{j \in \mathcal{P}_i} e^{s(i,j)}}{\sum_{k=1}^{B} e^{s(i,k)}},
\end{equation}
where $B$ is the batch size and $\mathcal{P}_i$ is the set of indices of point clouds positive to text $i$.
To ensure bidirectional alignment, the full contrastive loss $\mathcal{L}_{\text{cont}}$ is the symmetric average of this and the corresponding point-cloud-to-text loss $\mathcal{L}_{p \to t}$:
\begin{equation}
\mathcal{L}_{\text{cont}} = \frac{1}{2} (\mathcal{L}_{t \to p} + \mathcal{L}_{p \to t}).
\end{equation}

\begin{figure}[t]
    \centering
    \includegraphics[width=\columnwidth]{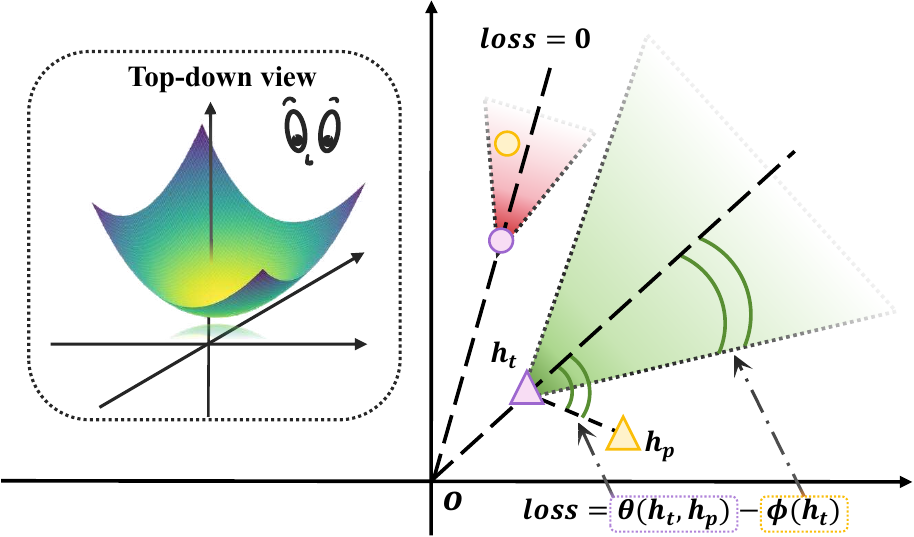}
    \caption{Geometric illustration of the Hierarchical Ordering Loss. The loss enforces the "text entails 3D" partial order by penalizing a 3D embedding $h_p$ only if it lies outside the entailment cone defined by its corresponding text embedding $h_t$. The penalty is proportional to the difference between the exterior angle $\theta(h_t, h_p)$ and the cone's half-aperture $\phi(h_t)$.}
    \label{fig:figure_3}
\end{figure}

\noindent \textbf{The Hierarchical Ordering Loss.} 
The hierarchical ordering loss $\mathcal{L}_{\text{ord}}$ mitigates HRC by embedding the “text entails 3D” partial order directly in hyperbolic geometry through entailment cones. For every text root embedding $\mathbf{h}_{t}\in\mathbb{H}^{d}_{c}$ we define a hyperbolic cone whose axis is $\mathbf{h}_{t}$ and whose half-aperture $\phi(\mathbf{h}_{t})$ contracts as the vector drifts outward from the origin, thereby capturing the intuition that concepts become more specific and semantically narrower with increasing radius. The half-aperture can be written as:
\begin{equation}
\phi(\vec{h}_t) = \operatorname{arcsin}\left(\frac{2K}{\sqrt{c} \|\tilde{\vec{h}}_t\|_E}\right),
\end{equation}
where $\|\tilde{\vec{h}}_t\|_E$ is the Euclidean norm of the spatial component of the text embedding, $c$ is the curvature, and $K=0.1$ caps the maximal cone width for concepts near the origin. Given a paired 3D embedding $\vec{h}_p$, we compute the exterior angle between the cone’s axis $\vec{h}_t$ and $\vec{h}_p$:
\begin{equation}
\theta(\vec{h}_t, \vec{h}_p) = \operatorname{arccos}\left(\frac{h_{p,d+1} + c \cdot h_{t,d+1} \langle \vec{h}_t, \vec{h}_p \rangle_{\mathcal{L}}}{\|\tilde{\vec{h}}_t\|_E \sqrt{(c \langle \vec{h}_t, \vec{h}_p \rangle_{\mathcal{L}})^2 - 1}}\right),
\end{equation}
where $h_{\cdot,d+1}$ denotes the time component and $\langle\cdot,\cdot\rangle_{\mathcal L}$ is the Lorentz inner product. The ordering loss penalises only those 3D points that fall outside the cone. As illustrated in Figure~\ref{fig:figure_3}, if the exterior angle exceeds the aperture, a penalty proportional to their difference is incurred. For a positive pair $(\vec{h}_t, \vec{h}_p)$, the loss is:
\begin{equation}
\mathcal{L}_{\text{ord}} = \max(0, \theta(\vec{h}_t, \vec{h}_p) - \phi(\vec{h}_t)).
\end{equation}

This nonsymmetric geometric constraint forces the text embedding to occupy a more general, "ancestor" position relative to its specific 3D instance, thereby constructing the desired hierarchy and preventing its collapse. The final training objective is a weighted sum of these two losses:
\begin{equation}
\mathcal{L}_{\text{total}} = \mathcal{L}_{\text{cont}} + \lambda \mathcal{L}_{\text{ord}},
\end{equation}
where the hyperparameter $\lambda$ balances discrimination against hierarchical consistency.

\begin{table*}[t]
\centering
\setlength{\tabcolsep}{0.5pt}{
\begin{tabular}{@{}cccccccccccccc@{}}
\toprule
\multirow{2}{*}[-0.7ex]{Datasets} & \multirow{2}{*}[-0.7ex]{Methods} & \multicolumn{2}{c}{Backbone} & \multicolumn{3}{c}{Text $\rightarrow$ PC} & \multicolumn{3}{c}{PC $\rightarrow$ Text} & \multirow{2}{*}[-0.7ex]{Rsum} & \multicolumn{3}{c}{Hyperparameters} \\ \cmidrule(lr){3-4} \cmidrule(lr){5-7} \cmidrule(lr){8-10} \cmidrule(lr){12-14}
& & Text & Point Cloud & R@1 & R@5 & R@10 & R@1 & R@5 & R@10 & & Batch Size & Nhead & SA Layers \\ \midrule
\multirow{5}{*}{T3DR-HIT} & RMARN$_{2025}$ & CLIP & PointNet & 13 & 39 & 47 & - & - & - & 99 & 64 & 16 & 6 \\
& RMARN$_{2025}$ & CLIP & PointNet & 19 & 50 & 53 & - & - & - & 122 & 32 & 32 & 6 \\
& RMARN$_{2025}$ & BERT & PointNet++ & 25 & 58 & 62 & - & - & - & 145 & 64 & 32 & 6 \\
& RMARN$_{2025}$ & BERT & PointNet++ & 31 & 61 & 69 & - & - & - & 161 & 64 & 32 & 8 \\
& \textbf{H$^2$ARN (Ours)} & \textbf{CLIP} & \textbf{DGCNN} & \textbf{32} & \textbf{63} & \textbf{73} & - & - & - & \textbf{168} & \textbf{256} & \textbf{64} & \textbf{6} \\ \midrule
\multirow{5}{*}{T3DR-HIT v2} & RMARN$_{2025}$ & CLIP & PointNet & 7.6 & 25.2 & 37.7 & 6.5 & 20.0 & 30.3 & 127.3 & 64 & 16 & 6 \\
& RMARN$_{2025}$ & CLIP & DGCNN & 13.4 & 38.3 & 58.3 & 18.4 & 40.9 & 51.0 & 220.3 & 32 & 32 & 8 \\
& H$^2$ARN (Ours) & CLIP & DGCNN & 15.6 & 43.3 & 58.6 & 15.0 & 37.2 & 55.2 & 224.9 & 256 & 16 & 6 \\
& H$^2$ARN (Ours) & CLIP & DGCNN & 14.7 & 42.9 & 59.5 & 18.5 & \textbf{44.7} & 54.4 & 234.7 & 32 & 32 & 8 \\
& H$^2$ARN (Ours) & CLIP & DGCNN & \textbf{16.9} & 44.4 & 59.4 & 16.4 & 41.0 & \textbf{56.9} & 235.0 & 256 & 32 & 8 \\
& \textbf{H$^2$ARN (Ours)} & \textbf{CLIP} & \textbf{DGCNN} & 16.4 & \textbf{44.5} & \textbf{60.6} & \textbf{19.6} & 42.3 & 55.1 & \textbf{238.5} & \textbf{256} & \textbf{64} & \textbf{6} \\ \bottomrule
\end{tabular}%
}
\caption{Performance comparison on the T3DR-HIT and our expanded T3DR-HIT v2 datasets.}
\label{tab:main_results}
\end{table*}
\section{Experiments} 
\subsection{Experimental Settings}
\noindent\textbf{Datasets.}
We validate our model using the original T3DR-HIT dataset as well as an expanded version, referred to as T3DR-HIT v2. The original dataset includes 3,380 text-3D pairs, but it presents a notable limitation: fine-grained artifact scenes are described by only a single caption per object, in contrast to indoor scenes, which typically include at least three captions. To correct this imbalance and evaluate scalability, we utilized the LLaVA large language model (\texttt{llava-v1.6-mistral-7b-hf}) to generate three additional, distinct captions for each artifact using a set of diverse prompts. This enhancement results in richer and more comprehensive textual representations. Concurrently, we also expanded the point cloud data by incorporating additional artifacts captured from the Elephant Meta Dataset provided by the Henan Broadcasting and Television Station. The augmented dataset, T3DR-HIT v2, contains a total of 8,935 text-3D pairs, representing a 2.6-fold increase in size. We partition the dataset into an 80:20 training and testing split and conduct experiments on both versions to thoroughly assess the effectiveness and robustness of our proposed model.

\noindent\textbf{Evaluation Metrics.}
To quantitatively evaluate retrieval performance, we employ two standard metrics widely used in cross-modal retrieval: Recall@K (R@K) and Rsum. R@K is defined as the proportion of queries for which the correct corresponding item is found within the top-K retrieved results. We report R@K for K=\{1, 5, 10\}, as this reflects performance at different levels of retrieval precision. To provide a single, comprehensive measure of overall performance, we also report Rsum, which is the sum of all R@K values across both retrieval directions (text-to-point cloud and point cloud-to-text). For all metrics, higher values signify better retrieval performance.

\noindent\textbf{Implementation Details.}
We implement our H$^{2}$ARN with the following architectural parameters and training settings. The shared latent dimension $d$ for all embeddings is set to 512, and the initial feature dimensions for both text ($D_t$) and point clouds ($D_p$) are also 512. The local feature sequence lengths are fixed at $L_t=77$ for text and $L_p=100$ for point clouds. Key parameters of the hyperbolic space are learnable. To ensure its positivity, the curvature parameter $c$ is parameterized via its logarithm, i.e., the model learns $\log(c)$, and is initialized to $c=1.0$. Similarly, to prevent numerical overflow during the hyperbolic projection, a modality-specific scaling factor $\alpha$ is applied to the local Euclidean features before aggregation. This factor is also learned via its logarithm and initialized as $\alpha=1/\sqrt{d}$.

We train the model for 100 epochs using a batch size of 256. For optimization, we employ the AdamW optimizer with a learning rate of $2 \times 10^{-3}$, and parameters $\beta_1=0.91$, $\beta_2=0.9993$, and $\epsilon=10^{-8}$. A linear learning rate scheduler with a warmup phase over the first 10\% of total training steps is used to stabilize training. For the dual geometric loss, the temperature $\tau$ in the contrastive loss is set to 0.07. The weight $\lambda$ for the hierarchical ordering Loss is set to 0.2, and the constant $K$ for the entailment cone is 0.1.

\subsection{Performance Comparison}
\begin{figure}[t]
    \centering
    \includegraphics[width=\columnwidth]{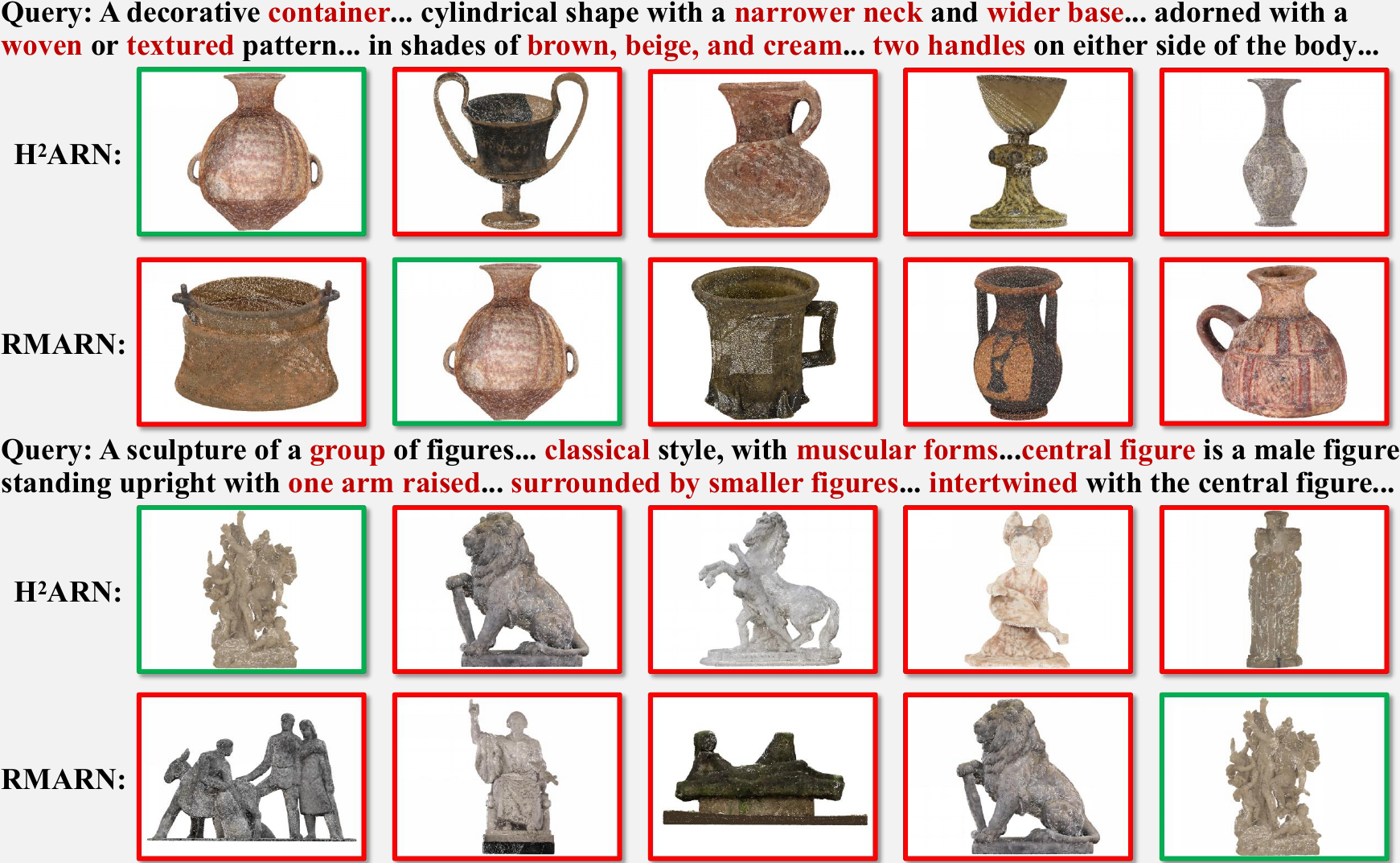}
    \caption{Qualitative comparison of text-to-3D retrieval results on the T3DR-HIT v2 dataset. For each query, the top-5 retrieved point clouds are shown, ranked from left to right by matching score. Green boxes indicate correct matches, while red boxes indicate incorrect ones.}
    \label{fig:qualitative_results}
\end{figure}
We compare H$^2$ARN with RMARN, the method that introduced the T3DR-HIT dataset and remains the only published baseline. On both the original and the expanded T3DR-HIT v2 datasets, H$^2$ARN consistently achieves superior quantitative performance, as shown in Table~\ref{tab:main_results}. On the original dataset, our model, equipped with CLIP and DGCNN backbones, sets a new state-of-the-art by outperforming RMARN across all R@K metrics for text-to-point cloud retrieval. This performance advantage becomes even more evident on the more challenging T3DR-HIT v2 dataset. When compared under identical backbone configurations, H$^2$ARN significantly outperforms RMARN in both retrieval directions, achieving an Rsum of 238.5 compared to RMARN’s 220.3. This marked improvement, independent of feature extractor choice, underscores the effectiveness of our key innovations: the use of hyperbolic geometry to address HRC and the introduction of contribution-aware aggregation to mitigate RISD. The results confirm our model's robustness and scalability to diverse scenarios, with qualitative examples in Figure~\ref{fig:qualitative_results} further demonstrating its accurate fine-grained retrieval.

\subsection{Ablation Study}
\begin{table}[t]
\centering
\setlength{\tabcolsep}{6pt}{\begin{tabular}{@{}cccccc@{}}
\toprule
\multirow{2}{*}[-0.7ex]{Methods} & \multicolumn{2}{c}{Text $\rightarrow$ PC} & \multicolumn{2}{c}{PC $\rightarrow$ Text} & \multirow{2}{*}[-0.7ex]{Rsum} \\ \cmidrule(lr){2-3} \cmidrule(lr){4-5}
& R@1 & R@5 & R@1 & R@5 & \\ \midrule
\textbf{H$^2$ARN (Ours)} & \textbf{16.4} & \textbf{44.5} & \textbf{19.6} & \textbf{42.3} & \textbf{238.5} \\
w/o $\mathcal{L}_{\text{ord}}$ & 15.3 & 40.1 & 18.4 & 41.1 & 229.6 \\
w/o Aggregation & 15.2 & 43.6 & 16.9 & 41.2 & 233.5 \\
w/o both & 14.3 & 41.8 & 14.5 & 37.5 & 222.0 \\
\bottomrule
\end{tabular}}
\caption{Ablation study on the core components of H$^2$ARN.}
\label{tab:ablation_components}
\end{table}

In this section, we conduct a series of ablation studies to analyze the impact of our key design choices.

\noindent\textbf{Effect of Core Components.}
We first validate the effectiveness of our two primary contributions: the Hierarchical Ordering Loss ($\mathcal{L}_{\text{ord}}$) and the Contribution-Aware Hyperbolic Aggregation mechanism. As shown in Table~\ref{tab:ablation_components}, removing the $\mathcal{L}_{\text{ord}}$ results in a significant performance drop, with the Rsum decreasing from 238.5 to 229.6. This decline highlights the crucial role of $\mathcal{L}_{\text{ord}}$ in explicitly modeling the semantic hierarchy and resolving the HRC problem. The impact of the Contribution-Aware Aggregation is also marked. When we ablate it by using the initial hyperbolic anchor (the mapped mean of leaf nodes) directly as the global representation, the model's performance degrades, with the Rsum falling to 233.5. This demonstrates that our aggregation mechanism is vital for filtering out redundant information and mitigating the RISD problem. Removing both components leads to a severe performance collapse to an Rsum of 222.0.

\noindent\textbf{Effect of Hyperbolic Space.}
\begin{table}[t]
\centering
\setlength{\tabcolsep}{6pt}{\begin{tabular}{@{}ccccccc@{}}
\toprule
\multirow{2}{*}[-0.7ex]{Methods} & \multicolumn{2}{c}{Text $\rightarrow$ PC} & \multicolumn{2}{c}{PC $\rightarrow$ Text} & \multirow{2}{*}[-0.7ex]{Rsum} \\ \cmidrule(lr){2-3} \cmidrule(lr){4-5}
& R@1 & R@5 & R@1 & R@5 & \\ \midrule
Eu + MP & 10.1 & 38.8 & 12.5 & 32.5 & 196.3 \\
Eu + CA & 12.5 & 41.0 & 14.2 & 36.4 & 215.1 \\
\textbf{H$^2$ARN (Ours)} & \textbf{16.4} & \textbf{44.5} & \textbf{19.6} & \textbf{42.3} & \textbf{238.5} \\
\bottomrule
\end{tabular}}
\caption{Ablation study on the effect of the hyperbolic space.}
\label{tab:ablation_space}
\end{table}
To verify the fundamental importance of hyperbolic geometry, we compare our full model against two Euclidean-based variants, with results in Table~\ref{tab:ablation_space}. Our first baseline, \textit{Eu + MP}, performs mean pooling on context-aware features and uses standard cosine similarity for alignment, yielding the lowest Rsum of 196.3. We then apply our contribution-aware aggregation within the Euclidean framework (\textit{Eu + CA}), where weights are determined by the dot product between leaves and their anchor. This variant significantly improves performance to an Rsum of 215.1, confirming that our aggregation mechanism is effective at mitigating RISD even without hyperbolic geometry. Finally, our full H$^2$ARN model, which leverages the hyperbolic embedding space and the hierarchical ordering loss, achieves the best performance by a large margin. This progression directly validates the superiority of hyperbolic geometry.

\noindent\textbf{Effect of Self-Attention Layers.}
We further analyze the impact of the number of attention heads and layers in the structural context encoder. The results, presented in Table~\ref{tab:ablation_sa}, reveal a complex interplay between model width and depth, with no single configuration dominating all individual metrics. The optimal settings for specific R@K metrics are distributed across different architectures. For instance, the best R@5 score for text-to-point cloud retrieval is achieved by a relatively shallow model (32 heads, 4 layers) at 46.1, whereas the point cloud-to-text direction favors a deeper, wider architecture (64 heads, 8 layers) with a score of 46.4. This indicates a clear trade-off between model width and depth depending on the specific evaluation criterion. Despite the varied performance on individual metrics, the configuration with 64 attention heads and 6 layers achieves the highest overall performance, reaching a peak Rsum of 238.5. 

\begin{table}[t]
\centering
\setlength{\tabcolsep}{10pt}{\begin{tabular}{@{}cccccc@{}}
\toprule
\multirow{2}{*}[-0.7ex]{\shortstack{(Nhead, \\ Layers)}} & \multicolumn{2}{c}{Text $\rightarrow$ PC} & \multicolumn{2}{c}{PC $\rightarrow$ Text} & \multirow{2}{*}[-0.7ex]{Rsum} \\ \cmidrule(lr){2-3} \cmidrule(lr){4-5}
& R@1 & R@5 & R@1 & R@5 & \\ \midrule
(16, 4) & 15.4 & 39.4 & 15.7 & 35.6 & 223.2 \\
(16, 6) & 15.6 & 43.3 & 15.0 & 37.2 & 224.9 \\
(16, 8) & 14.5 & 40.7 & 14.1 & 36.6 & 210.2 \\ \midrule
(32, 4) & 16.7 & \textbf{46.1} & 14.3 & 38.6 & 226.4 \\
(32, 6) & 16.5 & 43.2 & 14.4 & 41.7 & 231.0 \\
(32, 8) & 16.9 & 44.4 & 16.4 & 41.0 & 235.0 \\ \midrule
(64, 4) & 15.2 & 41.3 & \textbf{20.3} & 40.5 & 223.5 \\
(64, 6) & 16.4 & 44.5 & 19.6 & 42.3 & \textbf{238.5} \\
(64, 8) & \textbf{17.0} & 42.6 & 18.3 & \textbf{46.4} & 237.4 \\ \bottomrule
\end{tabular}}
\caption{Ablation study on the number of attention heads and self-attention layers on the T3DR-HIT v2 dataset.}
\label{tab:ablation_sa}
\end{table}

\section{Conclusion}
This paper introduces the Hyperbolic Hierarchical Alignment Reasoning Network (H$^2$ARN) for text-3D retrieval. By leveraging Lorentzian hyperbolic geometry, H$^2$ARN embeds both text and 3D data in a space naturally suited to hierarchical structures, enforcing semantic entailment through a geometric ordering loss. Its contribution-aware aggregation mechanism further enhances discriminative power by emphasizing semantically relevant features. Experimental results on both the original and expanded T3DR-HIT datasets confirm that H$^2$ARN significantly outperforms existing methods, demonstrating superior generalization and robustness. These contributions not only advance the state of the art in text-3D retrieval but also lay foundational groundwork for future cross-modal hyperbolic representations.

\section{Acknowledgments}
This work was supported in part by the National Key R\&D Program of China (2023YFA1008501) and the National Natural Science Foundation of China (NSFC) under grants 624B2049 and U22B2035.


\begin{thebibliography}{52}
\providecommand{\natexlab}[1]{#1}

\bibitem[{Bai et~al.(2025)Bai, Zhang, Zhao, Wu, Deng, Cui, Feng, and Xu}]{Bai02}
Bai, H.; Zhang, J.; Zhao, Z.; Wu, Y.; Deng, L.; Cui, Y.; Feng, T.; and Xu, S. 2025.
\newblock Task-driven Image Fusion with Learnable Fusion Loss.
\newblock In \emph{Proceedings of the IEEE/CVF Conference on Computer Vision and Pattern Recognition (CVPR)}, 7457--7468.

\bibitem[{Bai et~al.(2024)Bai, Zhao, Zhang, Wu, Deng, Cui, Jiang, and Xu}]{bai01}
Bai, H.; Zhao, Z.; Zhang, J.; Wu, Y.; Deng, L.; Cui, Y.; Jiang, B.; and Xu, S. 2024.
\newblock ReFusion: Learning Image Fusion from Reconstruction with Learnable Loss Via Meta-Learning.
\newblock \emph{International Journal of Computer Vision}, 1--21.

\bibitem[{Bao et~al.(2023)Bao, Wei, Zhou, Liu, Xie, Li, and Tian}]{bao2023multi}
Bao, L.; Wei, L.; Zhou, W.; Liu, L.; Xie, L.; Li, H.; and Tian, Q. 2023.
\newblock Multi-granularity matching transformer for text-based person search.
\newblock \emph{IEEE Transactions on Multimedia}, 26: 4281--4293.

\bibitem[{Bao et~al.(2025)Bao, Liu, Jiao, Liu, Li, Li, Liu, and Chen}]{bao02}
Bao, Q.; Liu, F.; Jiao, L.; Liu, Y.; Li, S.; Li, L.; Liu, X.; and Chen, P. 2025.
\newblock Visual-Language Scene-Relation-Aware Zero-Shot Captioner.
\newblock \emph{IEEE Transactions on Pattern Analysis and Machine Intelligence}, 47(10): 8725--8739.

\bibitem[{Bao et~al.(2022)Bao, Liu, Liu, Jiao, Liu, and Li}]{bao01}
Bao, Q.; Liu, F.; Liu, Y.; Jiao, L.; Liu, X.; and Li, L. 2022.
\newblock Hierarchical Scene Normality-Binding Modeling for Anomaly Detection in Surveillance Videos.
\newblock In \emph{Proceedings of the 30th ACM International Conference on Multimedia}, MM '22, 6103–6112. New York, NY, USA: Association for Computing Machinery.
\newblock ISBN 9781450392037.

\bibitem[{Chen et~al.(2018)Chen, Choy, Savva, Chang, Funkhouser, and Savarese}]{chen2018text2shape}
Chen, K.; Choy, C.~B.; Savva, M.; Chang, A.~X.; Funkhouser, T.; and Savarese, S. 2018.
\newblock Text2shape: Generating shapes from natural language by learning joint embeddings.
\newblock In \emph{Asian conference on computer vision}, 100--116. Springer.

\bibitem[{Chen et~al.(2024{\natexlab{a}})Chen, Huang, Xiong, and Lu}]{chen2024integrating}
Chen, Y.; Huang, J.; Xiong, S.; and Lu, X. 2024{\natexlab{a}}.
\newblock Integrating multisubspace joint learning with multilevel guidance for cross-modal retrieval of remote sensing images.
\newblock \emph{IEEE Transactions on Geoscience and Remote Sensing}, 62: 1--17.

\bibitem[{Chen et~al.(2024{\natexlab{b}})Chen, Lai, Chen, and Li}]{Chen_03}
Chen, Z.; Lai, Z.; Chen, J.; and Li, J. 2024{\natexlab{b}}.
\newblock Mind Marginal Non-Crack Regions: Clustering-Inspired Representation Learning for Crack Segmentation.
\newblock In \emph{Proceedings of the IEEE/CVF Conference on Computer Vision and Pattern Recognition (CVPR)}, 12698--12708.

\bibitem[{Chen et~al.(2022)Chen, Zhang, Lai, Chen, Liu, and Li}]{Chen_01}
Chen, Z.; Zhang, J.; Lai, Z.; Chen, J.; Liu, Z.; and Li, J. 2022.
\newblock Geometry-Aware Guided Loss for Deep Crack Recognition.
\newblock In \emph{Proceedings of the IEEE/CVF Conference on Computer Vision and Pattern Recognition (CVPR)}, 4703--4712.

\bibitem[{Chen et~al.(2023)Chen, Zhang, Lai, Zhu, Liu, Chen, and Li}]{Chen_02}
Chen, Z.; Zhang, J.; Lai, Z.; Zhu, G.; Liu, Z.; Chen, J.; and Li, J. 2023.
\newblock The Devil is in the Crack Orientation: A New Perspective for Crack Detection.
\newblock In \emph{Proceedings of the IEEE/CVF International Conference on Computer Vision (ICCV)}, 6653--6663.

\bibitem[{Diao et~al.(2021)Diao, Zhang, Ma, and Lu}]{diao2021similarity}
Diao, H.; Zhang, Y.; Ma, L.; and Lu, H. 2021.
\newblock Similarity reasoning and filtration for image-text matching.
\newblock In \emph{Proceedings of the AAAI conference on artificial intelligence}, volume~35, 1218--1226.

\bibitem[{Faghri et~al.(2017)Faghri, Fleet, Kiros, and Fidler}]{faghri2017vse++}
Faghri, F.; Fleet, D.~J.; Kiros, J.~R.; and Fidler, S. 2017.
\newblock Vse++: Improving visual-semantic embeddings with hard negatives.
\newblock \emph{arXiv preprint arXiv:1707.05612}.

\bibitem[{Fu et~al.(2023)Fu, Mao, Song, and Zhang}]{fu2023learning}
Fu, Z.; Mao, Z.; Song, Y.; and Zhang, Y. 2023.
\newblock Learning semantic relationship among instances for image-text matching.
\newblock In \emph{Proceedings of the IEEE/CVF Conference on Computer Vision and Pattern Recognition}, 15159--15168.

\bibitem[{Gromov(1987)}]{gromov1987hyperbolic}
Gromov, M. 1987.
\newblock Hyperbolic groups.
\newblock In \emph{Essays in group theory}, 75--263. Springer.

\bibitem[{Han et~al.(2019)Han, Shang, Wang, Liu, and Zwicker}]{han2019y2seq2seq}
Han, Z.; Shang, M.; Wang, X.; Liu, Y.-S.; and Zwicker, M. 2019.
\newblock Y2Seq2Seq: Cross-modal representation learning for 3D shape and text by joint reconstruction and prediction of view and word sequences.
\newblock In \emph{Proceedings of the AAAI Conference on Artificial Intelligence}, volume~33, 126--133.

\bibitem[{Lee(2018)}]{lee2018introduction}
Lee, J.~M. 2018.
\newblock \emph{Introduction to Riemannian manifolds}, volume~2.
\newblock Springer.

\bibitem[{Lee et~al.(2018)Lee, Chen, Hua, Hu, and He}]{lee2018stacked}
Lee, K.-H.; Chen, X.; Hua, G.; Hu, H.; and He, X. 2018.
\newblock Stacked cross attention for image-text matching.
\newblock In \emph{Proceedings of the European conference on computer vision (ECCV)}, 201--216.

\bibitem[{Li et~al.(2019)Li, Zhang, Li, Li, and Fu}]{li2019visual}
Li, K.; Zhang, Y.; Li, K.; Li, Y.; and Fu, Y. 2019.
\newblock Visual semantic reasoning for image-text matching.
\newblock In \emph{Proceedings of the IEEE/CVF international conference on computer vision}, 4654--4662.

\bibitem[{Li and Fan(2022)}]{li2022image}
Li, W.; and Fan, X. 2022.
\newblock Image-text alignment and retrieval using light-weight transformer.
\newblock In \emph{ICASSP 2022-2022 IEEE International Conference on Acoustics, Speech and Signal Processing (ICASSP)}, 4758--4762. IEEE.

\bibitem[{Li et~al.(2025{\natexlab{a}})Li, Han, Chen, Chai, Lu, Wang, and Fan}]{li2025riemann}
Li, W.; Han, W.; Chen, Y.; Chai, Y.; Lu, Y.; Wang, X.; and Fan, X. 2025{\natexlab{a}}.
\newblock Riemann-based multi-scale attention reasoning network for text-3D retrieval.
\newblock In \emph{Proceedings of the AAAI Conference on Artificial Intelligence}, volume~39, 18485--18493.

\bibitem[{Li et~al.(2025{\natexlab{b}})Li, Han, Deng, Xiong, and Fan}]{wenrui111}
Li, W.; Han, W.; Deng, L.-J.; Xiong, R.; and Fan, X. 2025{\natexlab{b}}.
\newblock Spiking Variational Graph Representation Inference for Video Summarization.
\newblock \emph{IEEE Transactions on Image Processing}, 34: 5697--5709.

\bibitem[{Li et~al.(2023{\natexlab{a}})Li, Ma, Deng, Fan, and Tian}]{wenrui333}
Li, W.; Ma, Z.; Deng, L.-J.; Fan, X.; and Tian, Y. 2023{\natexlab{a}}.
\newblock Neuron-Based Spiking Transmission and Reasoning Network for Robust Image-Text Retrieval.
\newblock \emph{IEEE Transactions on Circuits and Systems for Video Technology}, 33(7): 3516--3528.

\bibitem[{Li et~al.(2023{\natexlab{b}})Li, Ma, Shi, and Fan}]{li2023style}
Li, W.; Ma, Z.; Shi, J.; and Fan, X. 2023{\natexlab{b}}.
\newblock The style transformer with common knowledge optimization for image-text retrieval.
\newblock \emph{IEEE Signal Processing Letters}, 30: 1197--1201.

\bibitem[{Li et~al.(2025{\natexlab{c}})Li, Wang, Wang, Zuo, Fan, and Tian}]{wenrui444}
Li, W.; Wang, P.; Wang, X.; Zuo, W.; Fan, X.; and Tian, Y. 2025{\natexlab{c}}.
\newblock Multi-Timescale Motion-Decoupled Spiking Transformer for Audio-Visual Zero-Shot Learning.
\newblock \emph{IEEE Transactions on Circuits and Systems for Video Technology}, 35(11): 10772--10786.

\bibitem[{Li et~al.(2024)Li, Wang, Xiong, and Fan}]{wenrui222}
Li, W.; Wang, P.; Xiong, R.; and Fan, X. 2024.
\newblock Spiking Tucker Fusion Transformer for Audio-Visual Zero-Shot Learning.
\newblock \emph{IEEE Transactions on Image Processing}, 33: 4840--4852.

\bibitem[{Li, Xiong, and Fan(2024)}]{li2024multi}
Li, W.; Xiong, R.; and Fan, X. 2024.
\newblock Multi-layer probabilistic association reasoning network for image-text retrieval.
\newblock \emph{IEEE transactions on circuits and systems for video technology}, 34(10): 9706--9717.

\bibitem[{Li et~al.(2025{\natexlab{d}})Li, Yang, Han, Man, Wang, and Fan}]{yangzhe3}
Li, W.; Yang, Z.; Han, W.; Man, H.; Wang, X.; and Fan, X. 2025{\natexlab{d}}.
\newblock Hyperbolic-Constraint Point Cloud Reconstruction from Single RGB-D Images.
\newblock \emph{Proceedings of the AAAI Conference on Artificial Intelligence}, 39(5): 4959--4967.

\bibitem[{Li et~al.(2023{\natexlab{c}})Li, Ding, Chen, and Elhoseiny}]{li2023uni3dl}
Li, X.; Ding, J.; Chen, Z.; and Elhoseiny, M. 2023{\natexlab{c}}.
\newblock Uni3dl: Unified model for 3d and language understanding.
\newblock \emph{arXiv preprint arXiv:2312.03026}.

\bibitem[{Li et~al.(2025{\natexlab{e}})Li, Liao, Tang, Zhang, Li, Bian, Sheng, Feng, Li, Gao et~al.}]{zhuoyuan01}
Li, Z.; Liao, J.; Tang, C.; Zhang, H.; Li, Y.; Bian, Y.; Sheng, X.; Feng, X.; Li, Y.; Gao, C.; et~al. 2025{\natexlab{e}}.
\newblock USTC-TD: A test dataset and benchmark for image and video coding in 2020s.
\newblock \emph{IEEE Transactions on Multimedia}.

\bibitem[{Lin et~al.(2024)Lin, Cheng, Guo, Mao, and Li}]{lin2024sca}
Lin, D.; Cheng, Y.; Guo, A.; Mao, S.; and Li, Y. 2024.
\newblock SCA-PVNet: Self-and-cross attention based aggregation of point cloud and multi-view for 3D object retrieval.
\newblock \emph{Knowledge-Based Systems}, 296: 111920.

\bibitem[{Liu et~al.(2023)Liu, Zhang, Wang, Chen, Wang, Huang, Shen, and Wang}]{liu2023efficient}
Liu, C.; Zhang, Y.; Wang, H.; Chen, W.; Wang, F.; Huang, Y.; Shen, Y.-D.; and Wang, L. 2023.
\newblock Efficient token-guided image-text retrieval with consistent multimodal contrastive training.
\newblock \emph{IEEE Transactions on Image Processing}, 32: 3622--3633.

\bibitem[{Messina et~al.(2021{\natexlab{a}})Messina, Amato, Esuli, Falchi, Gennaro, and Marchand-Maillet}]{messina2021fine}
Messina, N.; Amato, G.; Esuli, A.; Falchi, F.; Gennaro, C.; and Marchand-Maillet, S. 2021{\natexlab{a}}.
\newblock Fine-grained visual textual alignment for cross-modal retrieval using transformer encoders.
\newblock \emph{ACM Transactions on Multimedia Computing, Communications, and Applications (TOMM)}, 17(4): 1--23.

\bibitem[{Messina et~al.(2021{\natexlab{b}})Messina, Falchi, Esuli, and Amato}]{messina2021transformer}
Messina, N.; Falchi, F.; Esuli, A.; and Amato, G. 2021{\natexlab{b}}.
\newblock Transformer reasoning network for image-text matching and retrieval.
\newblock In \emph{2020 25th International conference on pattern recognition (ICPR)}, 5222--5229. IEEE.

\bibitem[{Peng and Qi(2019)}]{peng2019cm}
Peng, Y.; and Qi, J. 2019.
\newblock CM-GANs: Cross-modal generative adversarial networks for common representation learning.
\newblock \emph{ACM Transactions on Multimedia Computing, Communications, and Applications (TOMM)}, 15(1): 1--24.

\bibitem[{Radford et~al.(2021)Radford, Kim, Hallacy, Ramesh, Goh, Agarwal, Sastry, Askell, Mishkin, Clark et~al.}]{radford2021learning}
Radford, A.; Kim, J.~W.; Hallacy, C.; Ramesh, A.; Goh, G.; Agarwal, S.; Sastry, G.; Askell, A.; Mishkin, P.; Clark, J.; et~al. 2021.
\newblock Learning transferable visual models from natural language supervision.
\newblock In \emph{International conference on machine learning}, 8748--8763. PmLR.

\bibitem[{Ruan et~al.(2024)Ruan, Lee, Zhang, Zhang, and Chang}]{ruan2024tricolo}
Ruan, Y.; Lee, H.-H.; Zhang, Y.; Zhang, K.; and Chang, A.~X. 2024.
\newblock Tricolo: Trimodal contrastive loss for text to shape retrieval.
\newblock In \emph{Proceedings of the IEEE/CVF Winter Conference on Applications of Computer Vision}, 5815--5825.

\bibitem[{Tang et~al.(2023)Tang, Yang, Wu, Han, and Chang}]{tang2023parts2words}
Tang, C.; Yang, X.; Wu, B.; Han, Z.; and Chang, Y. 2023.
\newblock Parts2words: Learning joint embedding of point clouds and texts by bidirectional matching between parts and words.
\newblock In \emph{Proceedings of the IEEE/CVF Conference on Computer Vision and Pattern Recognition}, 6884--6893.

\bibitem[{Tang et~al.(2025)Tang, Yan, Yin, Zhang, Zhang, Ma, Gao, and Jia}]{haocheng01}
Tang, H.; Yan, R.; Yin, X.; Zhang, Q.; Zhang, X.; Ma, S.; Gao, W.; and Jia, C. 2025.
\newblock HGC-Avatar: Hierarchical Gaussian Compression for Streamable Dynamic 3D Avatars.
\newblock \emph{arXiv preprint arXiv:2510.16463}.

\bibitem[{Wang et~al.(2025)Wang, Tian, Liang, Tian, and He}]{wang2025global}
Wang, D.; Tian, J.; Liang, X.; Tian, Y.; and He, L. 2025.
\newblock Global-aware Fragment Representation Aggregation Network for image--text retrieval.
\newblock \emph{Pattern Recognition}, 159: 111085.

\bibitem[{Wang et~al.(2019)Wang, Sun, Liu, Sarma, Bronstein, and Solomon}]{wang2019dynamic}
Wang, Y.; Sun, Y.; Liu, Z.; Sarma, S.~E.; Bronstein, M.~M.; and Solomon, J.~M. 2019.
\newblock Dynamic graph cnn for learning on point clouds.
\newblock \emph{ACM Transactions on Graphics (tog)}, 38(5): 1--12.

\bibitem[{Wei et~al.(2020)Wei, Zhang, Li, Zhang, and Wu}]{wei2020multi}
Wei, X.; Zhang, T.; Li, Y.; Zhang, Y.; and Wu, F. 2020.
\newblock Multi-modality cross attention network for image and sentence matching.
\newblock In \emph{Proceedings of the IEEE/CVF conference on computer vision and pattern recognition}, 10941--10950.

\bibitem[{Wu et~al.(2024)Wu, Li, Wang, and Xiong}]{wu2024com3d}
Wu, H.; Li, R.; Wang, H.; and Xiong, H. 2024.
\newblock COM3D: Leveraging Cross-View Correspondence and Cross-Modal Mining for 3D Retrieval.
\newblock In \emph{2024 IEEE International Conference on Multimedia and Expo (ICME)}, 1--6. IEEE.

\bibitem[{Wu et~al.(2019)Wu, Mao, Zhang, Jiang, Li, Sun, and Ma}]{wu2019unified}
Wu, H.; Mao, J.; Zhang, Y.; Jiang, Y.; Li, L.; Sun, W.; and Ma, W.-Y. 2019.
\newblock Unified visual-semantic embeddings: Bridging vision and language with structured meaning representations.
\newblock In \emph{Proceedings of the IEEE/CVF Conference on Computer Vision and Pattern Recognition}, 6609--6618.

\bibitem[{Xiao, Li, and Jia(2025)}]{xiao01}
Xiao, Z.; Li, Z.; and Jia, W. 2025.
\newblock Occlusion-Embedded Hybrid Transformer for Light Field Super-Resolution.
\newblock In \emph{Proceedings of the AAAI Conference on Artificial Intelligence}, volume~39, 8700--8708.

\bibitem[{Xiao and Wang(2025)}]{xiao02}
Xiao, Z.; and Wang, X. 2025.
\newblock Event-based Video Super-Resolution via State Space Models.
\newblock In \emph{Proceedings of the Computer Vision and Pattern Recognition Conference}, 12564--12574.

\bibitem[{Yang, Li, and Cheng(2025)}]{yangzhe1}
Yang, Z.; Li, W.; and Cheng, G. 2025.
\newblock SHMamba: Structured Hyperbolic State Space Model for Audio-Visual Question Answering.
\newblock \emph{IEEE Transactions on Audio, Speech and Language Processing}, 33: 3582--3593.

\bibitem[{Yang et~al.(2025)Yang, Li, Hou, and Cheng}]{yangzhe2}
Yang, Z.; Li, W.; Hou, J.; and Cheng, G. 2025.
\newblock Multi-modal spiking tensor regression network for audio-visual zero-shot learning.
\newblock \emph{Neurocomputing}, 629: 129636.

\bibitem[{Yu et~al.(2024)Yu, Wang, Li, Zhu, Liang, Wang, and Okumura}]{yu2024towards}
Yu, F.; Wang, Z.; Li, D.; Zhu, P.; Liang, X.; Wang, X.; and Okumura, M. 2024.
\newblock Towards cross-modal point cloud retrieval for indoor scenes.
\newblock In \emph{International Conference on Multimedia Modeling}, 89--102. Springer.

\bibitem[{Zhang et~al.(2020)Zhang, Lei, Zhang, and Li}]{zhang2020context}
Zhang, Q.; Lei, Z.; Zhang, Z.; and Li, S.~Z. 2020.
\newblock Context-aware attention network for image-text retrieval.
\newblock In \emph{Proceedings of the IEEE/CVF conference on computer vision and pattern recognition}, 3536--3545.

\bibitem[{Zhang et~al.(2025{\natexlab{a}})Zhang, Ma, Wang, Zhang, Zhang, and Zhang}]{zhang01}
Zhang, X.; Ma, J.; Wang, G.; Zhang, Q.; Zhang, H.; and Zhang, L. 2025{\natexlab{a}}.
\newblock Perceive-IR: Learning to Perceive Degradation Better for All-in-One Image Restoration.
\newblock \emph{IEEE Transactions on Image Processing}, 1--1.

\bibitem[{Zhang et~al.(2025{\natexlab{b}})Zhang, Zhang, Wang, Zhang, Zhang, and Du}]{zhang02}
Zhang, X.; Zhang, H.; Wang, G.; Zhang, Q.; Zhang, L.; and Du, B. 2025{\natexlab{b}}.
\newblock UniUIR: Considering Underwater Image Restoration as an All-in-One Learner.
\newblock \emph{IEEE Transactions on Image Processing}, 34: 6963--6977.

\bibitem[{Zheng et~al.(2020)Zheng, Zheng, Garrett, Yang, Xu, and Shen}]{zheng2020dual}
Zheng, Z.; Zheng, L.; Garrett, M.; Yang, Y.; Xu, M.; and Shen, Y.-D. 2020.
\newblock Dual-path convolutional image-text embeddings with instance loss.
\newblock \emph{ACM Transactions on Multimedia Computing, Communications, and Applications (TOMM)}, 16(2): 1--23.

\end{thebibliography}

\end{document}